# Pre-Processing-Free Gear Fault Diagnosis Using Small Datasets with Deep Convolutional Neural Network-Based Transfer Learning

Pei Cao
Graduate Research Assistant and Ph.D. Candidate
Department of Mechanical Engineering
University of Connecticut
191 Auditorium Road, Unit 3139
Storrs, CT 06269
USA

Shengli Zhang
Stanley Black & Decker, Global Tool & Storage Headquarter
Towson, MD 21286
USA

J. Tang[†]
Professor
Department of Mechanical Engineering
University of Connecticut
191 Auditorium Road, Unit 3139
Storrs, CT 06269
USA
Phone: (860) 486-5911, Email: jtang@engr.uconn.edu

Submitted to

[†] Corresponding author

# Pre-Processing-Free Gear Fault Diagnosis Using Small Datasets with Deep Convolutional Neural Network-Based Transfer Learning

[1]Pei Cao, [2]Shengli Zhang and [1]J. Tang[*]

[1]Department of Mechanical Engineering, University of Connecticut, Storrs, CT 06269, USA
[2]Stanley Black & Decker, Global Tool & Storage Headquarter, Towson, MD 21286

## Abstract

Early fault diagnosis in complex mechanical systems such as gearbox has always been a great challenge, even with the recent development in deep neural networks. The performance of a classic fault diagnosis system predominantly depends on the features extracted and the classifier subsequently applied. Although a large number of attempts have been made regarding feature extraction techniques, the methods require great human involvements are heavily depend on domain expertise and may thus be non-representative and biased from application to application. On the other hand, while the deep neural networks based approaches feature adaptive feature extractions and inherent classifications, they usually require a substantial set of training data and thus hinder their usage for engineering applications with limited training data such as gearbox fault diagnosis. This paper develops a deep convolutional neural network-based transfer learning approach that not only entertains pre-processing free adaptive feature extractions, but also requires only a small set of training data. The proposed transfer learning architecture consists of two parts; the first part is constructed with a piece of a pre-trained deep neural network that serves to extract the features automatically from the input, the second part is a fully connected stage to classify the features that needs to be trained using gear fault experiment data. The proposed approach performs gear fault diagnosis using pre-processing free raw accelerometer data and experiments with various sizes of training data were conducted. The superiority of the proposed approach is revealed by comparing the performance with other methods such as locally trained convolution neural network and angle-frequency analysis based support vector machine. The achieved accuracy indicates that the proposed approach is not only viable and robust, but also has the potential to be readily applicable to other fault diagnosis practices.

**Keywords:** gear fault diagnosis, deep convolution neural network, transfer learning

## 1. Introduction

In modern industry, the significance of condition monitoring and fault diagnosis has been ever-increasing due to the continuously raising standard for safety and quality. Gearbox, as one of the most common components used in contemporary machineries, is susceptible to failure under severe working conditions and thus requires the practice of fault diagnosis. Signal-based techniques have been shown to be an effective way to facilitate such practices (Kang et al, 2001; Randall, 2011; Marquez et al, 2012). As its name indicates, in a typical signal-based system, vibration signals are measured first, a feature extraction technique is then employed to characterize the fault-related features and a classifier is applied to predict fault occurrence in terms of type and severity. The accuracy of fault diagnosis is thus heavily correlated to the features extracted by the adopted technique.

Last decades have seen diverse attempts to identify and extract useful features from measured signals for fault diagnosis, which fall into three main categories: time-domain analysis (Zhou et al, 2008; Parey and Pachori, 2012), frequency domain-analysis (Fakhfakh et al, 2005; Li et al, 2015; Wen et al, 2015) and time-frequency analysis (Tang et al, 2010; Chaari et al, 2012; Yan et al, 2014, Zhang and Tang, 2018). Although decent results of fault diagnosis tasks have been reported, most of these studies indeed exploit the features of preference. In other words, the types of feature extracted are mostly based on domain expertise and manual decision that may not be the most sensitive ones to reflect the health condition of the machinery. Moreover, expertise-based feature extraction techniques require vast human involvements and design efforts. In certain mechanical systems, hand-crafted feature extraction techniques can surely recognize discriminative features by reducing noise. However, for complex systems, as faults occur primarily at materials level but their effects can only be observed indirectly at a system level, the measured signals are often complicated and overlaid by various unidentified components (Lu et al., 2012). Take gearbox systems for example, the measured vibration signals are composed of periodic meshing frequencies, their harmonics, random noise, and etc. The complex geometries of the gears and pinions in practical structures further perplex the situation. For such kind of systems, the aforementioned methods developed are considerably specific towards applications and working conditions.

---

[*] Corresponding author at: Department of Mechanical Engineering, University of Connecticut, Storrs, CT 06269, USA. Tel: (860)486-5911.
E-mail addresses: jiong.tang@uconn.edu (J. Tang).

To provide more generic solutions, more recently, deep neural network-based feature extraction techniques are progressively investigated, which aim to adaptively extract features and classify damage/fault with minimal tuning. For example, Zhang et al (2015) developed a deep learning network for degradation pattern classification and demonstrated the efficacy of the proposed model using turbofan engine dataset; Li et al (2016) proposed a deep random forest fusion technique to for gearbox fault diagnosis which achieves 97.68% classification accuracy; Weimer et al (2016) examined the usage of deep Convolutional Neural Network for industrial inspection and demonstrated excellent defect detection results; Ince et al (2017) developed a fast motor condition monitoring system using a 1-D Convolutional Neural Network which obtains a classification accuracy of 97.4%; and Abdeljaber et al (2017) performed real-time damage detection using Convolutional Neural Network and showcased satisfactory efficiency. Deep neural network is undoubtedly a powerful tool in condition monitoring, as an end-to-end hierarchical system, it inherently blends feature extraction and classification into a single adaptive learning frame. Nevertheless, the amount of training data required for satisfactory results depends on many aspects of the practice, such as the correctness of training samples, the number of classes, and the degree of separation between each class. Inevitably, the lack of labeled training samples because a common issue for most machinery fault analysis. To improve the performance given limited training data, researchers have combined pre-processing and data augmentation techniques, e.g., discrete wavelet transform (Saravanan and Ramachandran, 2010), antialiasing/decimation filter (Ince et al, 2017), and wavelet packet transform (Li et al, 2016), with neural networks for fault diagnosis. However, the pre-processing process may hurt the objective feature of neural networks, which is a dominant reason why neural networks are selected to facilitate health monitoring in the first place.

In this research, we present a deep neural network-based transfer learning approach utilizing limited time domain data for gearbox fault diagnosis. The proposed architecture is composed of two parts. Massive image data (1.2 million) from ImageNet are used first to train the original deep neural network model. Then the parameters of the original neural network are partially transferred to construct the first part of the proposed architecture. The second part accommodates the gear fault diagnosis task of interest and is further trained using experimentally generated gear fault data. The proposed approach entertains two major advantages compared to other contemporary methods. First, vibration signals from measurement can be used directly for fault diagnosis without pre-processing which is usually prone to bias. Second, the developed scheme is able to achieve great accuracy when training dataset is small. The rest of this paper is organized as follows. In Section 2, the background information of gear fault diagnosis and transfer learning is outlined. In Section 3, experimental studies are carried out using the proposed approach with uncertainties and noise; comparisons between different approaches are conducted as well. Finally, concluding remarks are summarized in Section 4.

## 2 Approach Formulations
### 2.1 Convolutional Neural Network (CNN)

Convolutional Neural Network (CNN) is a class of biologically inspired neural network featuring one or multiple convolutional layers that simulate human visual system (LeCun et al, 1990). In recent years, due to enhanced computing solutions and increased data size in various applications, CNN-based methods have shown significant improvements in performance and thus have become the most popular class of approaches for feature extraction tasks such as image classification (Krizhevsky et al, 2012), natural language processing (Kim, 2016), recommending systems (Van den Oord, et al, 2013) and fault detection (Ince et al, 2016). CNN learns how to extract and recognize characteristics of the target task by combining and stacking convolutional layers, pooling layers and fully connected layers in its architecture. Figure 1 demonstrates a simple CNN with a input layer to accept images input, a convolution layer to extract feature, a ReLU layer to augment features through non-linear transformation, a max pooling layer to reduce data size, and a fully connected layer combined with a softmax layer to classify the input to pre-defined labels. The parameters are trained through a training dataset and updated using back propagation algorithm to reflect the features of the task that may not be recognized otherwise. The basic mechanism of layers in CNN is outlined as follows.

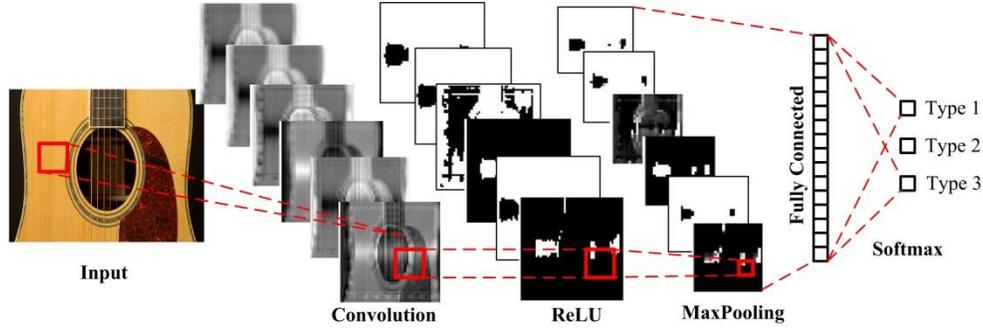
Figure 1 An example convolutional neural network

**Convolutional layer.** Each feature map in the convolution layer shown in Fig. 1 is generated by a convolution filter. Generally, if the input and convolution filters are tensors of size $m \times n$ and $p \times q \times K$ ($K$ is the number of filter used), respectively. Stride (step size of the filter sliding over input) is set to 1 and padding (how many rows and columns to insert around the original input) is set to 0. The convolution operation can be expressed as,

$$y_{d_1,d_2,k} = \sum_{i=0}^{p}\sum_{j=0}^{q} x_{d_1 i, d_2 j} \times f_{i,j,k} \quad (1)$$

where $y$, $x$ and $f$ denote the element in feature map, input and convolutional filter, respectively. $f_{i,j,k}$ represents the element on $i$-th column and $j$-th row for filter $k$. $y_{d_1,d_2,k}$ is the element on $d_1$-th column and $d_2$-th row of feature map $k$. And $x_{d_1 i, d_2 j}$ refers to the input element on $i$-th column and $j$-th row of the stride window specified by $d_1$ and $d_2$. Equation (1) gives a concise representation of the convolution operation when the input is 2D, and stride and padding are 1 and 0. Higher dimension convolution operations can be conducted in a similar manner. To be more evocative, suppose the input image can be represented by a $4 \times 7$ matrix and the convolution kernel is a $3 \times 3$ identity matrix. As we take kernel and stride it over the image matrix, dot products are taken in each step and recorded in a feature map matrix (Fig. 2). Such operation is called convolution. In CNN, multiple convolution filters are used in a convolutional layer, each acquires a feature piece in its own perspective from the input image specified by the filter parameters. Regardless of what and where a feature appears in the input, the convolution layer will try to characterize it from various perspectives that have been tuned automatically by the training dataset.

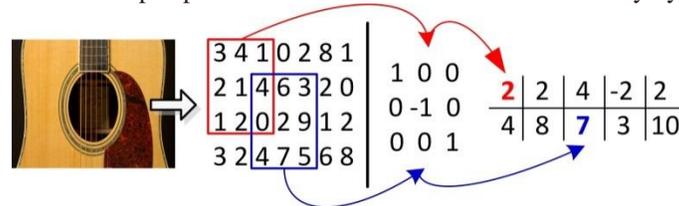
Figure 2 Illustration of convolution operation

**ReLU layer**. In CNN, ReLU (rectified linear units) layers are commonly used after convolution layers. Mostly, the relationship between the input and output is not linear. While the convolution operation is linear, the ReLU layer is designed to take non-linear relationship into account, as illustrated in the equation below,

$$\bar{y} = \max(0, y) \quad (2)$$

The ReLU operation is applied to each feature map and returns an activation map (Fig. 3). The depth of the ReLU layer equals to that of the convolution layer.

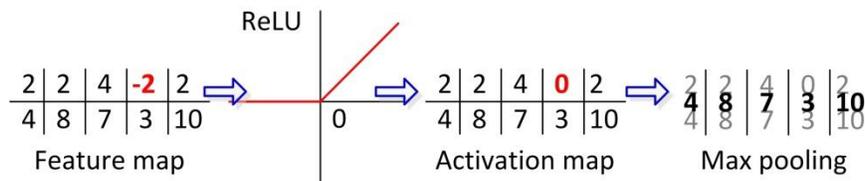
Figure 3 Illustration of ReLU and max pooling

**Max pooling layer**. Max pooling down samples a sub-region of the activation map to its maximum value,

$$\hat{y} = \max_{L_1 \leq i \leq U_1, L_2 \leq j \leq U_2} \overline{y}_{i,j} \quad (3)$$

where $L_1 \leq i \leq U_1, L_2 \leq j \leq U_2$ defines the sub-region. The max pooling layer not only makes the network less sensitive to location changes of a feature but also reduces the size of parameters, thus alleviating computational burden.

## 2.2 Transfer Learning

CNN is a powerful tool, and the performance can be further improved by upscaling the CNN equipped. However, the scale of a CNN concurs with the scale of the training dataset. Naturally, the deeper the CNN, the more parameters need to be trained, which requires a substantial amount of valid training samples. However, in applications toward certain mechanical systems, the training data is always not as sufficient as that of other tasks such as natural image classification.

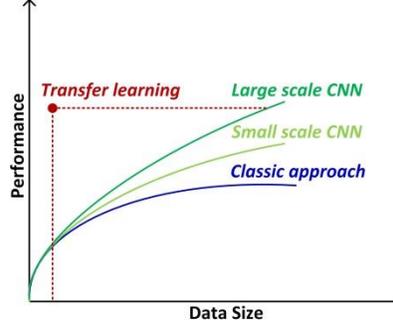

Figure 4 Learning methods: data size vs. performance

Figure 4 extracts the representative relationship between data size and performance for different learning methods. While the performance of a large scale CNN has the potential to top other methods, it is also profoundly correlated with the size of training data. Transfer learning, on the other hand, is able to achieve prominent performance commensurate with large scale CNNs using only a small set of training date. By partially deploying a pre-trained neural network from previous tasks, transfer learning provides a possible solution to improve the performance of a novel task with small training dataset. Classic transfer learning approaches transfer (copy) the first $n$ layers of a well-trained network to the target network of layer $m>n$. Initially, the last ($m-n$) layers of the target network are left untrained. They are trained subsequently using the training data from the novel task. Given training dataset from the previous task $\mathbf{D}_{pre}$ and the novel task $\mathbf{D}_{nov}$ as,

$$\mathbf{D}_{pre} = \{\mathbf{X}_{pre}, \mathbf{L}_{pre}\}$$
$$\mathbf{D}_{nov} = \{\mathbf{X}_{nov}, \mathbf{L}_{nov}\} \quad (4a, b)$$

where $\mathbf{X}$ is the input and $\mathbf{L}$ is the output label. The CNNs for both tasks can then be regarded as,

$$\hat{\mathbf{L}}_{pre} = \text{CNN}_{pre}(\mathbf{X}_{pre}, \boldsymbol{\theta}_{pre})$$
$$\hat{\mathbf{L}}_{nov} = \text{CNN}_{nov}(\mathbf{X}_{nov}, \boldsymbol{\theta}_{nov}) \quad (5a, b)$$

CNN denotes the mapping of a convolution neural network given parameters $\boldsymbol{\theta}$ from input to predicted output $\hat{\mathbf{L}}$. The parameters of the previous task is trained through,

$$\boldsymbol{\theta}_{pre}' = \arg\min_{\boldsymbol{\theta}_{pre}}(\mathbf{L}_{pre} - \hat{\mathbf{L}}_{pre}) = \arg\min_{\boldsymbol{\theta}_{pre}}(\mathbf{L}_{pre} - \text{CNN}_{pre}(\mathbf{X}_{pre}, \boldsymbol{\theta}_{pre})) \quad (6)$$

where $\boldsymbol{\theta}_{pre}'$ stands for the parameters after training. Thereupon, the trained parameters of the first $n$ layers can be transferred to the new task as,

$$\boldsymbol{\theta}_{nov}(1:n)' := \boldsymbol{\theta}_{pre}(1:n)' \quad (7)$$

The rest of the parameter can be trained using training samples from the novel task,

$$\boldsymbol{\theta}_{nov}(n:m)' = \arg\min_{\boldsymbol{\theta}_{nov}(n:m)}(\mathbf{L}_{nov} - \text{CNN}_{nov}(\mathbf{X}_{nov}, [\boldsymbol{\theta}_{nov}(1:n)', \boldsymbol{\theta}_{nov}(n:m)])) \quad (8)$$

Therefore, the CNN used for the novel task for future fault classification and diagnosis can be represented as,

$$\text{CNN}_{nov}(\mathbf{X}_{nov}, [\boldsymbol{\theta}_{nov}(1:n)', \boldsymbol{\theta}_{nov}(n:m)']) \quad (9)$$

where the parameters in first *n* layers are transferred from a previous task and the last (*m-n*) layers are trained using the training dataset of the novel task,

$$\boldsymbol{\theta}_{nov}' = [\boldsymbol{\theta}_{nov}(1:n)', \boldsymbol{\theta}_{nov}(n:m)'] \tag{10}$$

Transfer learning becomes possible and promising because, as has been discovered by some recent studies, the layers at the convolutional stages (convolution layers, ReLU layers and pooling layers) of the convolutional neural network trained on large dataset indeed extract general features of inputs, while the layers of fully connected stages (fully connected layers, softmax layers, classification layers) are more specific to task (Zeiler and Fergus, 2013; Sermanet et al, 2014). Therefore, the *n* layers transferred to the new task as a whole can be regarded as a well-trained feature extraction tool and the last few layers serve as a classifier to be trained. Even with substantial training data, initializing with transferred parameters can improve the performance in general (Yosinski et al, 2014).

In this research, transfer learning is implemented for gearbox fault diagnosis. The CNN is well-trained in terms of pulling characteristics from images. As illustrated in Fig. 5, the parameters in the convolution stage, i.e., the parameters used in the convolution filter, the ReLU operator and the max pooling operator are transferred to the fault diagnosis task. The parameters used in the fully connected layer and the softmax layers are trained subsequently using a small amount of training data generated from gear fault experiments. In the next sub-section, we present the details of the transfer learning architecture adopted in this study.

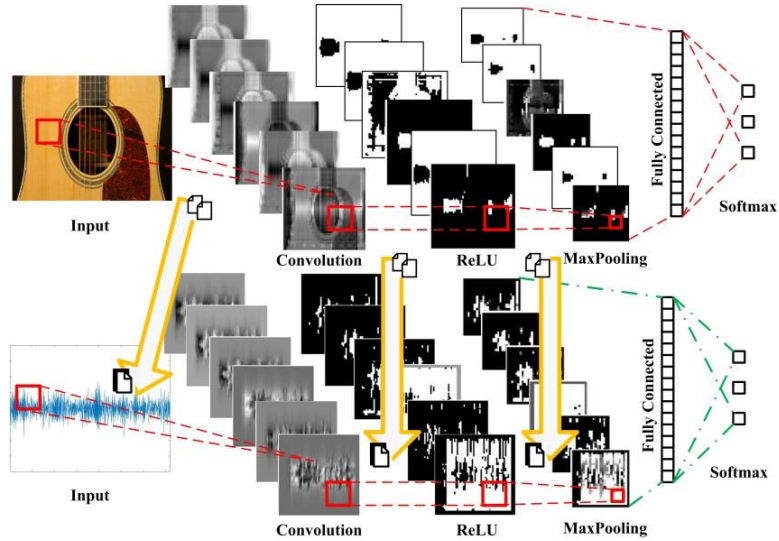

Figure 5 Illustration of transfer learning

### 2.3 Proposed Architecture

The deep CNN adopted in this study as base architecture is proposed in (Krizhevsky et al, 2012), which is essentially composed of 5 convolution stages and 3 fully connected stages (Figure 6). Such architecture is employed due to its extraordinary performance in Large Scale Visual Recognition Challenge 2012 (ILSVRC2012) and has been since repurposed for other learning tasks (Shie et al, 2015). To the best of our knowledge, the architecture has yet to be used for fault diagnosis using time domain inputs.

In the base architecture, the parameters are trained using roughly 1.2 million human/software labeled 3D true-color nature images from ImageNet. The trained parameters in the first 7 stages are well-polished in the sense of characterizing high-level abstractions of the input image and thus have the potential to be used for other tasks with image inputs. In gear fault diagnosis, vibration signals can be sampled using accelerometers as gear rotates. Such vibration signals can then be represented by 3D grey-scale/true-color images as shown in Fig. 5 and later in Fig. 9. Although the vibration images may look different from the images used to train the original CNN, useful features can be extracted in a similar manner as long as the CNN adopted is able to identify high-level abstractions. Therefore, as a deep convolution neural network, the first 7 stages of the base architecture can be transferred to facilitate gear fault diagnosis. As discussed in Section 2.2, the first 7 stages indeed serve as a general well-trained tool for automatic feature extraction. The more stages and layers used, the higher level of features can be obtained. The final stage is left to be trained as a classifier using the experiment data specific to the fault diagnosis task. As specified in Table 1, a total number of 24 layers are used in the proposed architecture; the parameters and specifications used in the first 21 layers are transferred from the base architecture.

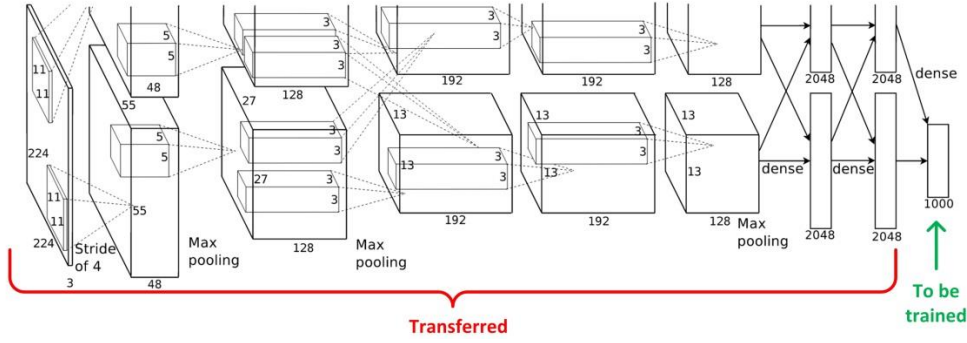
Figure 6 Illustration of the transfer learning architecture (adopted from (Krizhevsky et al, 2012))

Table 1 Specifications of the proposed architecture

| Stage | Layer | Name | Specifications |
|---|---|---|---|
| **1** (transferred) | 1 | Convolution | 11*11*96 |
|  | 2 | ReLU | N/A |
|  | 3 | Normalization | 5 channels/element |
|  | 4 | Max pooling | 3*3 |
| **2** (transferred) | 5 | Convolution | 5*5*256 |
|  | 6 | ReLU | N/A |
|  | 7 | Normalization | 5 channels/element |
|  | 8 | Max pooling | 3*3 |
| **3** (transferred) | 9 | Convolution | 3*3*384 |
|  | 10 | ReLU | N/A |
| **4** (transferred) | 11 | Convolution | 3*3*384 |
|  | 12 | ReLU | N/A |
| **5** (transferred) | 13 | Convolution | 3*3*256 |
|  | 14 | ReLU | N/A |
|  | 15 | Max pooling | 3*3 |
| **6** (transferred) | 16 | Fully connected | 4096 |
|  | 17 | ReLU | N/A |
|  | 18 | Dropout | 50% |
| **7** (transferred) | 19 | Fully connected | 4096 |
|  | 20 | ReLU | N/A |
|  | 21 | Dropout | 50% |
| **8 (to be trained)** | 22 | Fully connected | 9 |
|  | 23 | Softmax | N/A |
|  | 24 | Classification | Cross entropy |

In this study, the loss function used is the cross entropy function given as follows,

$$E(\mathbf{\theta}) = -\hat{\mathbf{L}} \cdot \ln(CNN(\mathbf{X}, \mathbf{\theta})) + \gamma \|\mathbf{\theta}\|_2 = -\hat{\mathbf{L}} \cdot \ln \mathbf{L} + \gamma \|\mathbf{\theta}\|_2 \quad (11)$$

where $\|\mathbf{\theta}\|_2$ is a $l_2$ normalization term to prevent the network from over-fitting. Equation (11) quantifies the difference between correct output labels and predicted labels. And the loss is then back propagated to update the parameters using the stochastic gradient descent method (Sutskever et al, 2013) given as,

$$\mathbf{\theta}_{i+1} = \mathbf{\theta}_i - \alpha \nabla E(\mathbf{\theta}_i) + \beta(\mathbf{\theta}_i - \mathbf{\theta}_{i-1}) \quad (12)$$

where $\alpha$ is the learning rate, $i$ is the number of iteration, and $\beta$ stands for the contribution of previous gradient step. The transferability of the base architecture and the performance of the proposed architecture for gear fault diagnosis will be investigated in the next section.

## 3. Experimental Studies
### 3.1 Gearbox Dynamics Experimental Setup

Many types of mechanical faults and failures can occur to gears in a gearbox. Vibration signals collected from such a system are usually used to reveal information about its operating condition. In this study, experimental cases are carried out on a two-stage gearbox with replaceable gears as shown in Fig. 7. The speed of the gear is controlled by a motor. The torque is supplied by a magnetic brake which can be adjusted by changing its input voltage. A 32-tooth pinion and an 80-tooth gear are installed on the first stage input shaft. The second stage consists of a 48-tooth pinion and 64-tooth gear. The input shaft speed is measured by a tachometer and gear vibration signals are measured by an accelerometer. The signals are recorded through a dSPACE system (DS1006 processor board, dSPACE Inc., Wixom, MI) with sampling frequency of 20 KHz. As illustrated in Fig. 8, nine different gear faults are introduced to the pinion on the input shaft including health, missing tooth, root crack, spalling and chipping tip with 5 different levels of severity. Dynamic responses of a system involving gear mechanism are angle-periodic. The gearbox system is usually treated as time-periodic system while the rotational speed is assumed to be constant. This assumption is generally not accurate because of load disturbances, geometric tolerances, and motor control errors, etc (Zhang and Tang, 2018). In this study, the originally time-domain vibration signals are converted from time-even to angle-even with evenly angular increment.

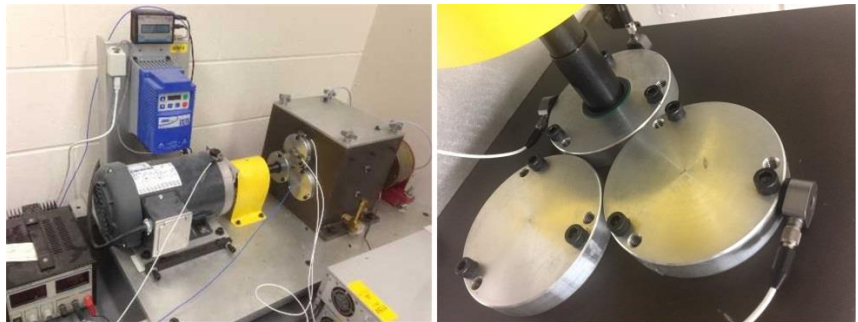

Figure 7 Gearbox for experimental study

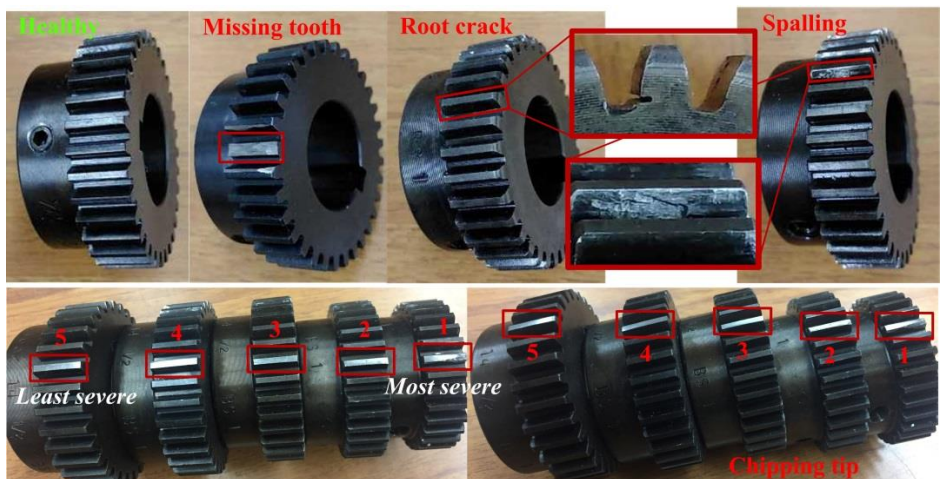

Figure 8 Nine pinions with different health conditions (five levels of severity for chipping tip)

For each gear condition, 104 signals are generated using the gearbox system. For each signal, 3,600 angle-even samples are recorded in the course of 4 gear revolutions first for the case study in Section 3.3, and then down sampled to 900 angle-even point for the case study in Section 3.4. Figure 9 shows 20 example signals of each type of gear condition where the vertical axis is the acceleration of the gear (rad/s$^2$) and the horizontal axis corresponds to the 3,600 angel-even sampling points.

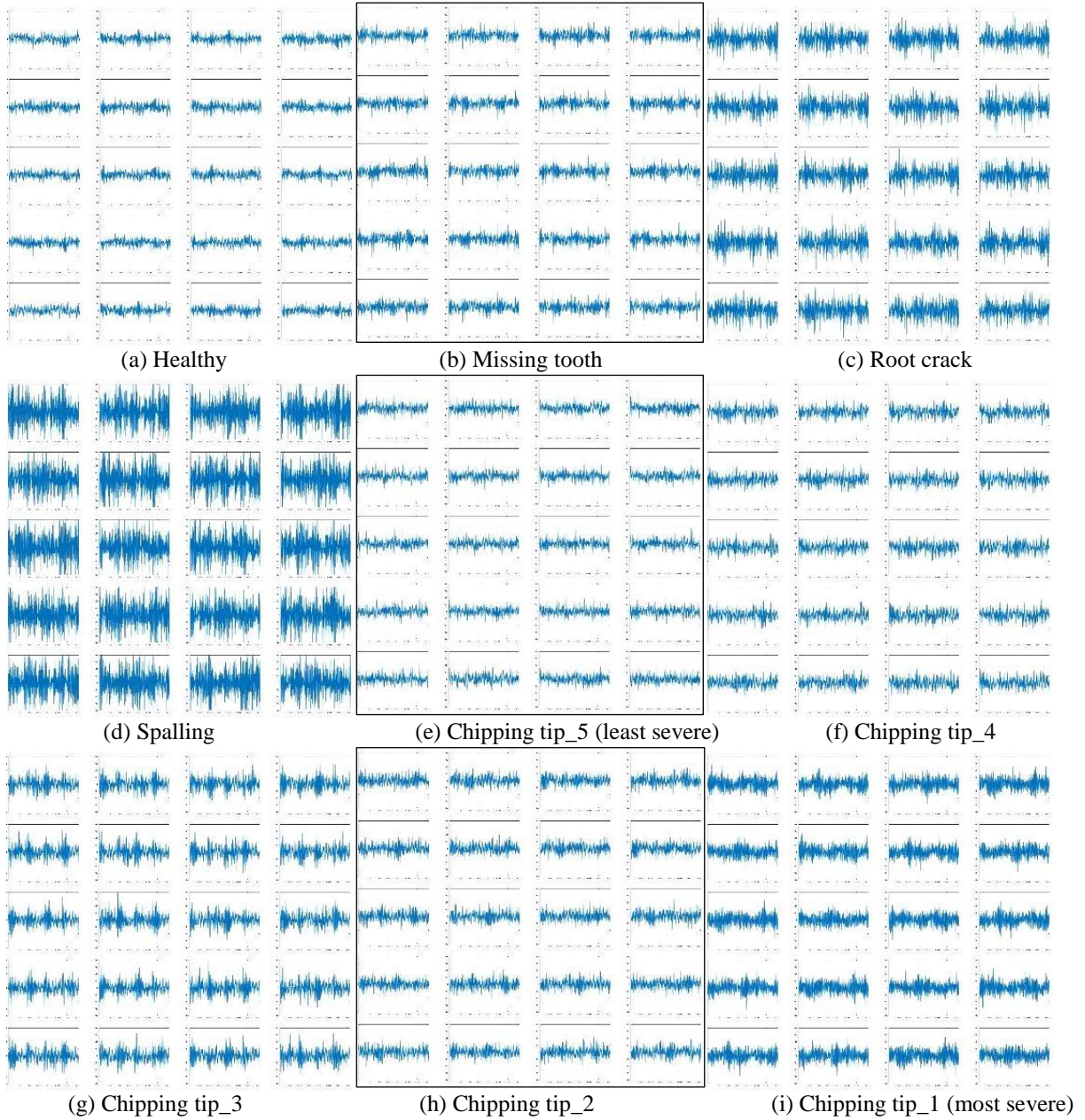

Figure 9 Vibration signal samples of different gear health conditions

(a) Healthy (b) Missing tooth (c) Root crack (d) Spalling (e) Chipping tip_5 (least severe) (f) Chipping tip_4 (g) Chipping tip_3 (h) Chipping tip_2 (i) Chipping tip_1 (most severe)

### 3.2 Network Setup and Methodology Comparison

In this study, the proposed transfer learning approach is examined and compared with two other approaches. In the first approach, a CNN consists of two convolution stages and a fully connected stage is employed, thereafter referred to as local CNN. The local CNN will be merely trained by data locally generated from gearbox experiments. The specifications are the same as the stage 1, stage 2 and stage 8 given in Table 1. The second approach adopted for comparison is angle-frequency domain synchronous analysis (AFS) (Zhang and Tang, 2018) followed by Support Vector Machine (SVM) classification. For the proposed transfer learning approach and the local-trained CNN approach, mini-batch size is set to 5, and 15 epochs are conducted meaning the training datasets are used for training 15 times throughout. The learning rate $\alpha$ is set to 1e-4 and 1e-2 for transferred layers and non-transferred layers, respectively. The momentum $\beta$ in Equation (12) is set to 0.9 for transfer learning and 0.5 for local CNN. For SVM approach, Gaussian kernel is adopted. In the next sub-sections, the relative performance of the three approaches is highlighted as we change the sampling frequency as well as the size of the training dataset.

### 3.3 Case 1 – 3,600 sampling points with varying training data size

As mentioned in Section 3.1, 104 vibration signals are generated for each gear condition. In the case studies, a portion of the signals are randomly selected as training data while the rest serves as validation data. To demonstrate the performance of the proposed approach towards various data sizes, the size of the training dataset ranges from 80% (83 training data per condition, 83*9 data in total) to 2% (2 training data per condition, 2*9 data in total) of all the 104 signals for each health condition.

Table 2 Classification results (3,600 sampling points)

| Method<br>Training data | Transfer learning Accuracy (%) | | Local CNN Accuracy (%) | | AFS-SVM Accuracy (%) | |
|---|---|---|---|---|---|---|
| **80%**<br>**(83 per condition)** | 100<br>100<br>100<br>100<br>100<br>100 | Mean:<br>100 | 91.01<br>99.47<br>97.35<br>100<br>100 | Mean:<br>97.57 | 86.72<br>88.62<br>87.80<br>87.26<br>86.99 | Mean:<br>87.48 |
| **60%**<br>**(62 per condition)** | 100<br>100<br>100<br>100<br>100 | Mean:<br>100 | 90.48<br>97.62<br>58.99<br>88.89<br>67.72 | Mean:<br>80.74 | 87.30<br>87.83<br>88.62<br>87.04<br>87.83 | Mean:<br>87.72 |
| **40%**<br>**(42 per condition)** | 100<br>100<br>100<br>100<br>100 | Mean:<br>100 | 88.89<br>98.39<br>44.44<br>62.72<br>83.69 | Mean:<br>76.63 | 86.74<br>86.38<br>85.84<br>87.99<br>86.38 | Mean:<br>86.67 |
| **20%**<br>**(21 per condition)** | 100<br>100<br>100<br>99.60<br>100 | Mean:<br>99.92 | 61.31<br>72.56<br>85.41<br>70.41<br>58.77 | Mean:<br>69.69 | 86.48<br>86.08<br>85.01<br>86.35<br>87.28 | Mean:<br>86.24 |
| **10%**<br>**(10 per condition)** | 99.88<br>98.23<br>99.88<br>99.29<br>99.76 | Mean:<br>99.41 | 64.07<br>57.09<br>55.56<br>44.56<br>57.80 | Mean:<br>55.82 | 80.97<br>86.17<br>78.84<br>86.29<br>86.88 | Mean:<br>83.83 |
| **5%**<br>**(5 per condition)** | 99.55<br>97.19<br>80.02<br>98.09<br>99.66 | Mean:<br>94.90 | 65.54<br>37.71<br>31.99<br>28.17<br>57.13 | Mean:<br>44.11 | 75.31<br>84.85<br>81.14<br>73.29<br>84.85 | Mean:<br>79.89 |
| **2%**<br>**(2 per condition)** | 76.80<br>73.31<br>69.39<br>73.42<br>68.19 | Mean:<br>72.22 | 26.14<br>27.67<br>32.24<br>31.70<br>22.22 | Mean:<br>27.99 | 61.87<br>73.97<br>41.72<br>69.72<br>64.92 | Mean:<br>62.44 |

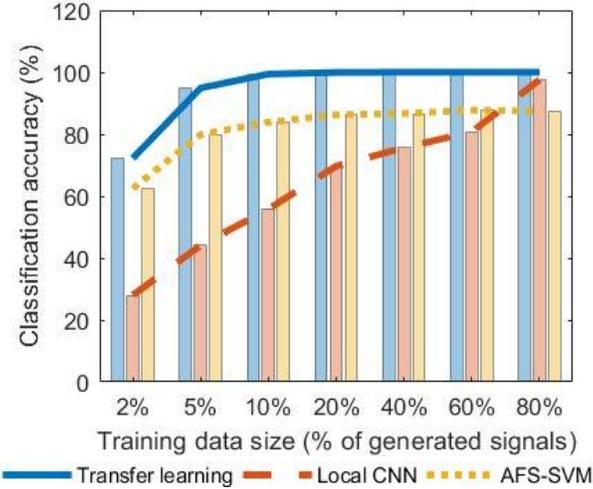
Figure 10 Classification results of the three methods when training data size varies

Table 2 shows the classification results where the mean accuracy is the average of 5 training attempts. The classification accuracy is the ratio of the correctly classified validation data to the total validation dataset. As illustrated in Fig. 10, the proposed transfer learning approach has the best classification accuracy for all types of data size. Even if only 5 vibration signals per condition are selected for training, the proposed approach is able to achieve an extraordinary 94.90% classification accuracy, which further escalates to 99%-100% when 10% and more training data are used. On the other hand, while the performance of AFS-SVM reaches the plateau (showing only minimal increments) after 20% date is used for training, the classification accuracy of local CNN gradually increases with data size from 27.99% to 97.57% and surpasses AFS-SVM eventually when 80% data is used for training, indicating the significance of the size of training data in order to properly train a neural network. Although the data size greatly affect the performance of a CNN in general senses, the proposed transfer learning architecture still exhibits high classification accuracy because only one fully connected stage needs to be trained locally, which notably lowers the standard of the data required by a CNN in terms of achieving satisfactory outcome. Figure 11 shows the convergent histories (mini-batch accuracy and mini-batch loss) of the proposed approach and local CNN when 5% data is used as the training dataset. As can be seen from the comparisons, transfer learning gradually converges in terms of both accuracy and loss as the training iterates while local CNN inclines to 'random walk' due to insufficient data. Compared with AFS-SVM, the proposed approach not only excels in performance, but also requires no pre-processing effort, angel-frequency analysis in this case, which makes the proposed approach more unbiased in feature extraction and readily applicable to other fault diagnosis practices. The proposed approach also shows satisfactory outcomes it the regard of robustness. As demonstrated in Fig. 12, it has the smallest variance in all cases, while the performance of the under-trained local CNN oscillates the most.

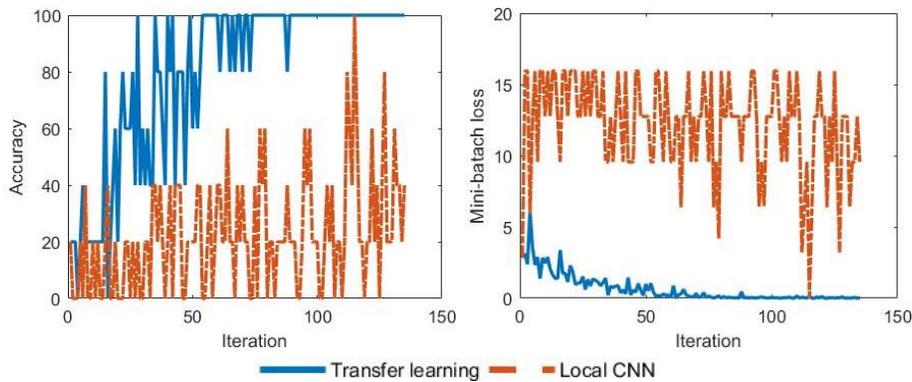
Figure 11 Convergent histories of transfer learning and local CNN for 5% training data

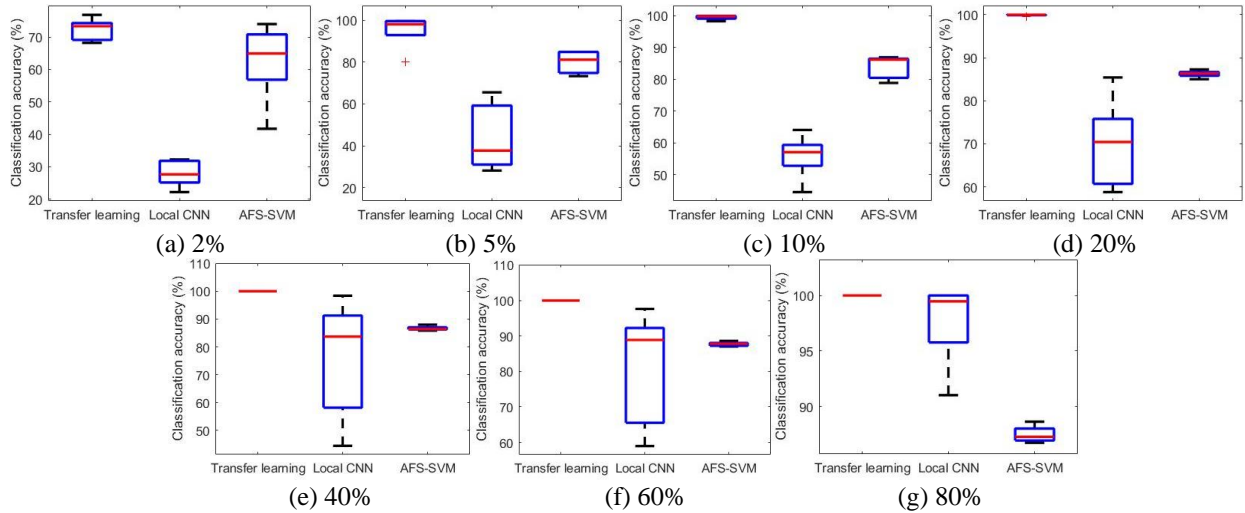

(a) 2%  (b) 5%  (c) 10%  (d) 20%

(e) 40%  (f) 60%  (g) 80%

Figure 12 Box plots of classification results of the three methods when training data size varies

As discussed in Section 2.2 and Section 2.3, the transferred stages of the proposed architecture attend to extract the high-level abstract features of the input that cannot be recognized otherwise, even if the input is different from that of the previous task. Figure 13 gives an example of such procedure by showing the feature maps generated in each convolution layer by the proposed architecture when it is used to classify a gearbox vibration signal. It is seen that the abstraction level of the input image continuously escalates from the 1$^{st}$ feature map to the 5$^{th}$ feature map. In general, the number of convolution stages equipped is correlated with the level of abstraction the features can be represented in the CNN. As demonstrated in this case study, the base architecture is indeed transferable towards gear fault diagnosis tasks and the proposed approach performs well with raw image signal inputs, which indicates the transferred layers constructed in this study are generally applicable to represent useful features of an input image in high-level abstraction.

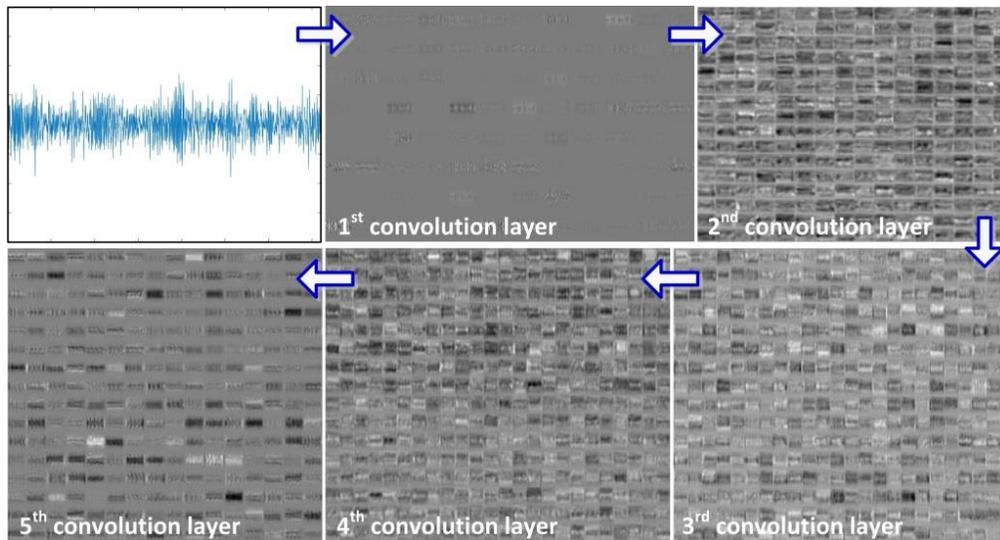

Figure 13 Feature maps extracted by 5 convolution layers of the proposed transfer learning approach

**3.4 Case 2 – 900 sampling points with varying training data size**

In Case 1, each vibration signal is composed of 3,600 angel-even accelerometer data points in the course of 4 gear revolutions. In some gear fault diagnosis systems, however, the sampling rate is lower than that of our first experiment. To take such systems into consideration, we down sampled the original vibration signals to 900 angel-even data points (Fig. 14) and apply the three methods for classification. Table 3 demonstrates the classification results of the three methods for different training data sizes. Similar to Case 1, the proposed transfer learning approach is the best performer for all scenarios. Figure 15 illustrates the classification results before and after down sampling. Although lowering sampling rate deteriorates the overall performance of all approaches, each method exhibits the similar trend as seen in Section 3.3. For transfer learning, it starts with 60.11% classification accuracy and reaches 95.88% when only 20% of data is used as training data whilst the accuracy of local CNN and AFS-SVM are 43.56% and 70.07%. Local CNN performs better than AFS-SVM when 80% data is used for training. Unlike AFS-SVM, the performance of local CNN can be largely improved if significantly more training data is incorporated because the parameters of lower stages can be learned from scratch. Eventually, the performance of the local CNN could reach the standard of the proposed transfer learning approach. Nevertheless, for cases with limited data, the proposed transfer learning approach is still the best option which has an extensive performance margin compared to local CNN or other pre-processing-based shallow learning methods such as AFS-SVM. Even with ample training data, initializing with transferred parameters can still improve the classification accuracy in general. And best of all, the proposed approach requires no pre-processing. Similarly to case 1 in Section 3.3, the proposed approach is still very robust especially when 40% or more data is used for training (Fig. 16).

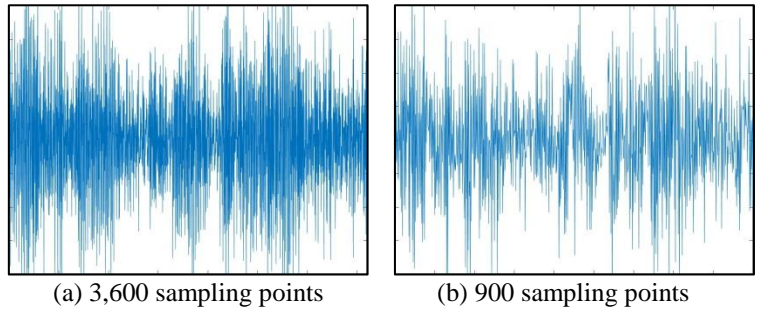

(a) 3,600 sampling points (b) 900 sampling points
Figure 14 Vibration signal of a spalling gear

Table 3 Classification results (900 sampling points)

| Method / Training data | Transfer learning Accuracy (%) | | Local CNN Accuracy (%) | | AFS-SVM Accuracy (%) | |
|---|---|---|---|---|---|---|
| **80% (83 per condition)** | 100<br>100<br>100<br>100<br>100 | Mean: 100 | 85.26<br>65.66<br>71.32<br>80.89<br>92.53 | Mean: 79.13 | 74.07<br>74.60<br>75.13<br>76.72<br>74.07 | Mean: 74.92 |
| **60% (62 per condition)** | 100<br>99.21<br>99.74<br>98.68<br>99.47 | Mean: 99.42 | 77.25<br>57.67<br>63.76<br>72.22<br>48.41 | Mean: 63.86 | 75.40<br>74.34<br>71.16<br>74.07<br>70.90 | Mean: 73.17 |
| **40% (42 per condition)** | 99.10<br>99.10<br>98.92<br>98.92<br>96.77 | Mean: 98.56 | 62.90<br>74.91<br>56.63<br>38.35<br>62.72 | Mean: 59.10 | 74.19<br>72.94<br>72.58<br>73.66<br>72.22 | Mean: 73.12 |
| **20% (21 per condition)** | 94.91<br>95.72<br>92.77<br>98.80<br>97.19 | Mean: 95.88 | 34.27<br>40.56<br>44.44<br>44.71<br>53.82 | Mean: 43.56 | 70.15<br>72.69<br>68.41<br>69.21<br>69.88 | Mean: 70.07 |
| **10% (10 per condition)** | 94.68<br>93.38<br>90.07<br>92.08<br>95.98 | Mean: 93.24 | 27.78<br>39.83<br>46.57<br>17.97<br>41.37 | Mean: 34.70 | 68.20<br>68.68<br>65.96<br>66.78<br>65.96 | Mean: 67.12 |
| **5%** | 70.73 | Mean: | 24.88 | Mean: | 64.42 | Mean: |
| **(5 per condition)** | 86.65 | 84.83 | 15.80 | 28.27 | 62.96 | 63.43 |
| | 89.12 | | 23.20 | | 63.86 | |
| | 90.24 | | 33.40 | | 65.66 | |
| | 87.43 | | 44.05 | | 60.27 | |
| **2%** <br> **(2 per condition)** | 55.99 <br> 59.91 <br> 58.06 <br> 61.11 <br> 65.47 | Mean: <br> 60.11 | 21.90 <br> 17.43 <br> 22.22 <br> 9.59 <br> 11.11 | Mean: <br> 16.45 | 47.93 <br> 57.30 <br> 58.06 <br> 51.53 <br> 55.12 | Mean: <br> 53.99 |

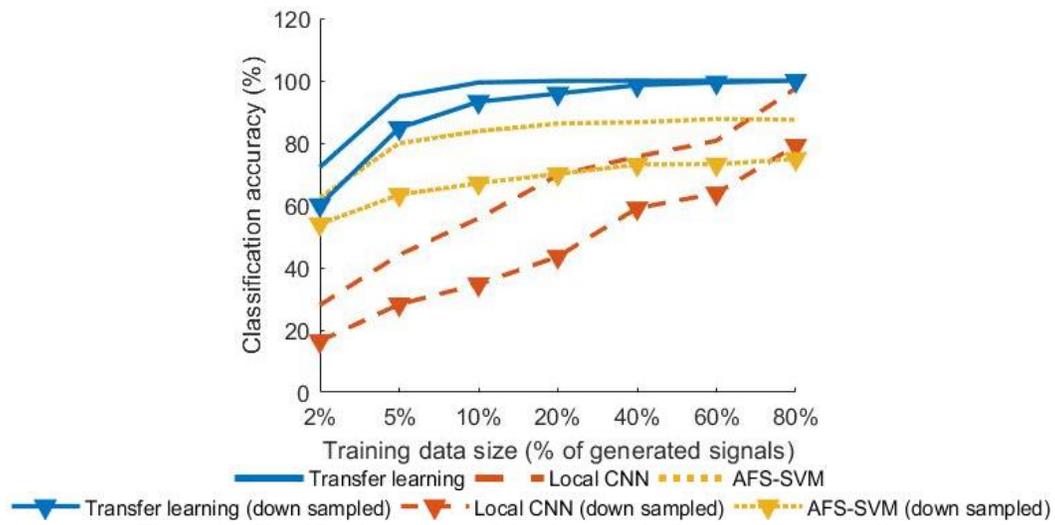

Figure 15 Classification results of the three methods after down sampling

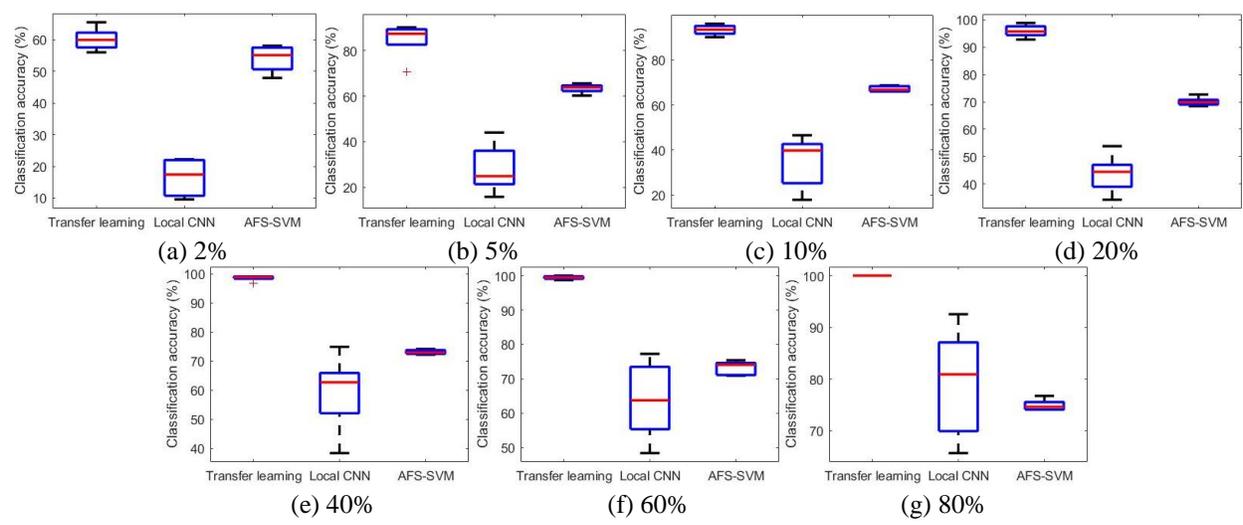

(a) 2%    (b) 5%    (c) 10%    (d) 20%

(e) 40%    (f) 60%    (g) 80%

Figure 16 Box plots of classification results of the three methods after down sampling

### 4. Concluding Remarks
In this paper, a deep convolutional neural network-based transfer learning approach is developed for deep feature extraction. The proposed approach not only entertains pre-processing free adaptive feature extractions, but also requires only a small set of training data compared to other locally trained convolution neural networks. The proposed transfer learning architecture consists of two parts; the first part is constructed with a piece of a pre-trained deep neural network that serves to extract the features automatically from the input, the second part is a fully connected stage for classification which needs to be trained using experiment data. Experiment studies have been conducted using pre-processing free raw accelerometer data towards gear fault diagnose. The performance of the proposed approach is highlighted by varying the size of training data. The classification accuracies outperform other methods such as locally trained convolution neural network and angle-frequency analysis-based support vector machine by as much as 50%. The achieved outcome indicates that the proposed approach is not only remarkable in gear fault diagnosis, but also has the potential to be readily applicable to other fault diagnosis practices.